\documentclass{ieeeaccess}
\usepackage{cite}
\usepackage{amsmath,amssymb,amsfonts}
\usepackage{algorithmic}
\usepackage{graphicx}
\usepackage{textcomp}

\usepackage{bm}

\usepackage{textcomp}
\usepackage{color}
\usepackage{multirow}
\usepackage{booktabs}
\usepackage{amsmath}
\usepackage{accents}
\usepackage{kotex}
\usepackage{hyperref}
\usepackage{color}

\makeatletter
\AtBeginDocument{\DeclareMathVersion{bold}
\SetSymbolFont{operators}{bold}{T1}{times}{b}{n}
\SetSymbolFont{NewLetters}{bold}{T1}{times}{b}{it}
\SetMathAlphabet{\mathrm}{bold}{T1}{times}{b}{n}
\SetMathAlphabet{\mathit}{bold}{T1}{times}{b}{it}
\SetMathAlphabet{\mathbf}{bold}{T1}{times}{b}{n}
\SetMathAlphabet{\mathtt}{bold}{OT1}{pcr}{b}{n}
\SetSymbolFont{symbols}{bold}{OMS}{cmsy}{b}{n}
\renewcommand\boldmath{\@nomath\boldmath\mathversion{bold}}}
\makeatother

\def\BibTeX{{\rm B\kern-.05em{\sc i\kern-.025em b}\kern-.08em
    T\kern-.1667em\lower.7ex\hbox{E}\kern-.125emX}}

\begin{document}
\history{Date of publication xxxx 00, 0000, date of current version xxxx 00, 0000.}
\doi{10.1109/ACCESS.2024.0429000}

\title{BioBridge: Unified Bio-Embedding with Bridging Modality in Code-Switched EMR.}
\author{
\uppercase{Jangyeong Jeon}\authorrefmark{1}, \IEEEmembership{Graduate Student Member, IEEE},
\uppercase{Sangyeon Cho}\authorrefmark{1},
\uppercase{Dongjoon Lee}\authorrefmark{1},
\uppercase{Changhee Lee}\authorrefmark{2},
and \uppercase{Junyeong Kim}\authorrefmark{1} \IEEEmembership{Member, IEEE}
}
\address[1]{Department of Artificial Intelligence, Chung-Ang University, Seoul 06974, Republic of Korea}
\address[2]{Department of Artificial Intelligence, Korea University, Seoul 02841, Republic of Korea}

\tfootnote{This work was partly supported by Institute of Information \& communications Technology Planning \& Evaluation (IITP) grant funded by the Korea government(MSIT) (No.2022-0-00184, Development and Study of AI Technologies to Inexpensively Conform to Evolving Policy on Ethics) and partly supported by Institute of Information \& communications Technology Planning \& Evaluation (IITP) grant funded by the Korea government(MSIT) (No.2021-0-01341, Artificial Intelligence Graduate School Program, Chung-Ang University)}

\markboth
{J. Jeon \headeretal: BioBridge: Unified Bio-Embedding with Bridging Modality in Code-Switched EMR.}
{J. Jeon \headeretal: BioBridge: Unified Bio-Embedding with Bridging Modality in Code-Switched EMR.}

\corresp{Corresponding author: Junyeong Kim (e-mail: junyeongkim@cau.ac.kr).}

\begin{abstract}
Pediatric Emergency Department (PED) overcrowding presents a significant global challenge, prompting the need for efficient solutions. 
This paper introduces the BioBridge framework, a novel approach that applies Natural Language Processing (NLP) to Electronic Medical Records (EMRs) in written free-text form to enhance decision-making in PED. 
In non-English speaking countries, such as South Korea, EMR data is often written in a Code-Switching(CS) format that mixes the native language with English, with most code-switched English words having clinical significance. 
The BioBridge framework consists of two core modules: ``bridging modality in context'' and ``unified bio-embedding.''
The “bridging modality in context” module improves the contextual understanding of bilingual and code-switched EMRs. 
In the “unified bio-embedding” module, the knowledge of the model trained in the medical domain is injected into the encoder-based model to bridge the gap between the medical and general domains. 
Experimental results demonstrate that the proposed BioBridge significantly performance traditional machine learning and pre-trained encoder-based models on several metrics, including F1 score, area under the receiver operating characteristic curve (AUROC), area under the precision-recall Curve (AUPRC), and Brier score. 
Specifically, BioBridge-{XLM} achieved enhancements of 0.85\% in F1 score, 0.75\% in AUROC, and 0.76\% in AUPRC, along with a notable 3.04\% decrease in the Brier score, demonstrating marked improvements in accuracy, reliability, and prediction calibration over the baseline XLM model. The source code will be made publicly available.
\footnote{Source code available at \url{https://github.com/jjy961228/BioBridge}}
\end{abstract}

\begin{keywords}
Natural Language Processing, Code-Switching, Electronic Medical Record, Emergency Department, Pediatric Emergency Department
\end{keywords}

\titlepgskip=-21pt

\maketitle

\section{Introduction}
\PARstart{O}{vercrowding} in Emergency Departments (EDs), including Pediatric Emergency Departments (PEDs), is an urgent problem that needs to be handled globally and immediately, including in South Korea. 
It is due to lead inefficient use of ED resources and the overworked healthcare professionals\cite{bucci2016emergency,zhou2018time,rasouli2019outcomes,sartini2022overcrowding}.
Providing timely and accurate treatment is essential, especially for children who are more susceptible to infection than adults\cite{american2004overcrowding}.
However, since children tend to visit the ED frequently, this further aggravates the overcrowding problem in PED\cite{kim2018understanding}.
%

As medical institutions worldwide adopt EMRs, vast amounts of healthcare data are stored digitally\cite{williams2008role,turner2017word2vec,hartswood2003making,menachemi2011benefits}.
EMR contains information about a patient's Present Illness (PI), past medical history, and textual data such as treatment and disposition information\cite{hartswood2003making,williams2008role,menachemi2011benefits}.
By using Natural Language Processing (NLP) techniques to EMR data, clinicians can receive real-time decision support and increase the efficiency of emergency medical services by optimizing the assessment of patient severity and urgency in the ED\cite{sterling2019prediction,yang2022large,mermin2023use}.
Traditional approaches to applying NLP to EMR data have focused on Machine Learning (ML)\cite{segura2018predicting,rivas2018automatic,alzoubi2018automated} and Word2Vec-based approaches to extract context-free word embeddings\cite{mikolov2013efficient,choi2016multi,zhang2019biowordvec}.
Recently, the focus has shifted to improving predictive modeling performance by training EMRs on pre-trained encoder-based models\cite{yu2019biobert,rasmy2021med, alsentzer2019publicly} such as BERT\cite{devlin2018bert}.
Despite the potential benefits of EMRs, their practical application in healthcare decision support systems has been limited by challenges in applying NLP to extract key information, which still needs to be overcome \cite{polnaszek2016overcoming,perera2013challenges,adnan2020role}.

First, EMRs consist of unstructured, free-text clinical notes written by physicians during patient examinations, often using varied wording to describe the same disease\cite{harrison2021machine,zhang2019high,castro2015validation}.
For example, medical symbols such as ``Cough/Sputum/Rhinorrhea'' are often used as ``C/S/R'' in EMRs, which are primarily used to record a patient's respiratory disease\cite{park2017current,shinozaki2020electronic}.
Extensive text preprocessing is required to apply natural language processing techniques to these EMRs, and NLP toolkits such as the Unified Medical Language System (UMLS)\cite{cimino2003consistency,bodenreider2004unified} have been widely adopted\cite{lindberg1993unified,cimino2003consistency,morrey2009neighborhood}.
However, these standardized preprocessing methods incur high costs due to continuous resource maintenance and are often unavailable in languages other than English, including Korean.
To mitigate these challenges, we have identified commonly used medical symbols and abbreviations in EMRs, proposed a straightforward preprocessing method to decode them, and integrated this preprocessed text with an ML-based\cite{chen2016xgboost,cox1958regression,friedman2001greedy,chen2016xgboost,breiman2001random} and pre-trained encoder-based model\cite{KoBERT,lee2020kr,devlin2018bert,conneau2019cross,conneau-etal-2020-unsupervised}.

Second, EMR data used in non-English speaking countries are recorded in Code-Switching (CS) form, where English and the native language are mixed \cite{kim2020korean,bae2021keyword}. 
CS refers to using two languages in the sentence and is prevalent not only in EMRs but also in multicultural countries, social media, and online platforms \cite{auer2013code,baker2011foundations,ahn2017language}.
Unfortunately, there is a lack of research on CS in non-English speaking countries, including languages such as Korean, Japanese, Chinese, and Arabic\cite{ahn2017language,park1990korean,miwa1985intrasentential,amazouz2017addressing}.
In particular, few studies have explored the application of encoder-based models to code-switched EMR datasets from non-English-speaking countries \cite{li2020chinese,kim2020korean,liu2021use,bae2021keyword}.
The EMR dataset used in our study is also code-switched to Korean-English.
To bridge this research gap, we assume that language-specific modalities (Korean, English) exist in each CS sentence and propose a novel method of ``bridging modality in the context'' of learning each modality effectively by a pre-trained encoder.
This approach is motivated by research in multimodal learning, where different inputs (visual, text, and audio input) are defined as modalities in a transformer and utilized for learning\cite{li2021bridging}.
In the following Section \ref{subsec : Multimodal Learning}, we describe in detail the research related to multimodal learning.

Third, most of the terms that are code-switched have clinical significance.
Using pre-trained encoder base models in general domains in NLP and applying them to clinical notes is challenging, and many attempts have been made to solve it \cite{beltagy2019scibert,yu2019biobert,rasmy2021med}.
This is due to most encoder-based models are not pre-trained with domain-specific data but are pre-trained using general-domain data such as English Wikipedia, BooksCorpus\cite{zhu2015aligning}, and CommonCrawl\cite{wenzek2019ccnet}. 
Moreover, Korean is a very low-resourced language compared to English\cite{lee2020kr,kim2022pre}, and more research needs to be done. 
Pre-trained encoder-based models on general domains have strengths in understanding general contexts but tend to perform poorly in medical domains\cite{yu2019biobert,rasmy2021med,kim2022pre}. 
To mitigate this gap, we propose a ``unified bio-embedding'' approach that effectively fine-tunes pre-trained encoder-based models\cite{KoBERT,lee2020kr,devlin2018bert,conneau2019cross,conneau-etal-2020-unsupervised}, bridging the gap between general and medical domains.
%

Our goal is to alleviate overcrowding in PEDs.
To achieve this, we propose to fine-tune the pre-trained encoder-based model to classify emergency and non-emergency cases\cite{fuchs2016definitions} using EMR for the three to overcome the abovementioned challenges. 
We introduce ``unified-bio embedding with bridging modality in code-switched EMR (BioBridge),'' a framework comprising two training modules: (1) bridging modality in context, and (2) unified bio-embedding.
We established Machine Learning (ML) and Transformer encoder-based models as baselines to validate BioBridge. 
The experimental results show that BioBridge has a surprisingly stable performance in predicting emergency cases. 
This is expected to alleviate PED overcrowding and is the first attempt to our knowledge. 
Our contributions are summarized as follows:
\begin{itemize}
    \item{Using a straightforward preprocessing method, we empirically identify and decode frequently used medical symbols and abbreviations in EMRs.}
    \item{We propose a method for effectively learning code-switched sentences.}
    \item{We propose a method to bridge the gap between general and medical domains, enhancing fine-tuning applicability.}
\end{itemize}

\section{Related Work}
\subsection{Natural Language Processing For Electronic Medical Record}
The increasing availability of Electronic Medical Records (EMRs) has led to extensive research into utilizing EMRs in Natural Language Processing (NLP)\cite{segura2018predicting,reddy2018predicting,rasmy2021med,yao2021novel}.
Traditional methods for applying EMR data to NLP have evolved from Machine Learning (ML)\cite{cox1958regression,friedman2001greedy,chen2016xgboost,breiman2001random} and Word2Vec-based approaches\cite{mikolov2013efficient}.
Word2Vec-based approaches have proven effective in identifying relationships between medical terms and symptoms by facilitating word encoding within a vector space where semantically similar words are closely aligned\cite{mikolov2013efficient,choi2016multi}. 
However, Word2Vec is limited by its inability to account for context, and methodologies such as Sent2Vec\cite{pagliardini2017unsupervised} have attempted to overcome this limitation by extending the Continuous Bag of Words (CBOW) approach of Word2Vec to increase the window size to whole sentences.
In particular, BioSent2Vec\cite{chen2019biosentvec}, trained on over 30 million documents from academic articles in PubMed and clinical notes in the MIMMIC-III Clinical Database, has been able to generate useful word embeddings through biomedical text mining. 
In this paper, we use the pre-trained BioSent2Vec as a word-level feature extractor to derive English medical word embeddings.

Encoder-based models such as BERT\cite{devlin2018bert} are gaining prominence in modern NLP due to their capacity to comprehend bidirectional context.
Specifically, such as BioBERT\cite{yu2019biobert} have been effectively fine-tuned on English EMR data for patient health prediction\cite{alsentzer2019publicly,rasmy2021med}.
However, the effectiveness of these encoder-based models is often constrained when applied to languages with limited resources, such as Korean\cite{lee2020kr,bae2021keyword,gao2021named,kim2022pre}. 
This limitation stems from the fact that English, an alphabet-based language, consists of only 26 characters and consequently faces fewer out-of-vocabulary (OOV) problems\cite{lee2020kr}. 
In contrast, Korean is a syllable-based language comprising 11,172 characters\cite{lee2020kr}.
Korean medical texts, in particular, pose analytical challenges due to complex expressions and specialized jargon\cite{kim2022pre,kim2020korean,bae2021keyword}.
EMRs in Korea frequently involve CS between Korean and English and are typically maintained in a free-text format\cite{castro2015validation,zhang2019high,harrison2021machine}.
To address these challenges, approaches have been proposed to fine-tune multilingual-specific encoder base models\cite{devlin2018bert,conneau2019cross,conneau2019unsupervised} such as multilingual BERT (mBERT)\cite{devlin2018bert} and Korean-specific encoder base models\cite{lee2020kr,KoBERT} such as KR-BERT\cite{lee2020kr} with EMR data. 
Notable studies include one focused on extracting clinically significant information from free-text clinical notes\cite{kim2023identifying}, another on effectively triaging emergency patients\cite{lee2024deep}, and another on de-identifying radiology reports\cite{an2023identification}.

The EMRs used for training in these studies cannot be a universal method for learning EMRs because the data cannot be released due to they contain personal patient information\cite{kim2023identifying,an2023identification,an2023identification}. 
Moreover, they only propose fine-tuning with EMR data without special techniques.
In addition, previous studies have reported results using only one type of model, either multilingual-only or Korean-only encoder base models, which may not fully capture the nuances of EMRs documented in both Korean and English.

In this paper, we conduct extensive experiments with both model types to address these limitations. We also introduce an approach to effectively learn from universally code-switched EMRs without disclosing patient data.

\subsection{Multimodal Learning}
\label{subsec : Multimodal Learning}
Initial research in multimodal learning primarily focused on integrating text and images\cite{lu2019vilbert,tan2019lxmert}. 
Subsequently, video and text representations could provide complementary information, and research on combining video and text representations became a significant focus.
For example, VideoBERT\cite{sun2019videobert,zhou2020unified} pursued richer information representation by combining video clips and text. 
Recently, advancements in Visual Question Answering (VQA) tasks in multimodal learning have centered on enhancing the capabilities of models to respond to complex questions across different modality types\cite{li2021bridging,gao2019dynamic,kim2019progressive,kim2020modality}.
These multimodal learning methods share a common goal of solving vision-and-language transfer tasks by separating inputs from different modalities (visual, text, audio) and integrating them into a transformer-based model.

Motivated by the methodology in multimodal learning, this paper hypothesizes the presence of distinct modalities for each language in also code-switched sentences involving two languages.

 A notable example is the proposed method presented in\cite{li2021bridging}, which introduced a straightforward yet effective method for tackling the Audio-Visual Scene-Aware Dialog (AVSD) task. 
 The AVSD task is to analyze a video clip to identify information about its events, characters, and environment and to provide appropriate answers when the user asks questions about them.
 The AVSD task is given video, audio, caption, and dialog history as input. 
 The study in\cite{li2021bridging} proposed an approach to concatenate video-audio, caption, and dialog history into one long sequence feature and add segment tokens for “[video],” “[caption],” “[user1]”, and “[user2]” to distinguish each integrated modality to understand each modality better in AVSD task. 

 Inspired by the methodology in\cite{li2021bridging}, this paper hypothesizes the presence of distinct modalities for each language in also code-switched sentences involving two languages.

\section{Proposed Method}

\begin{figure*}[ht!]
\begin{center}
\includegraphics[width=\textwidth]{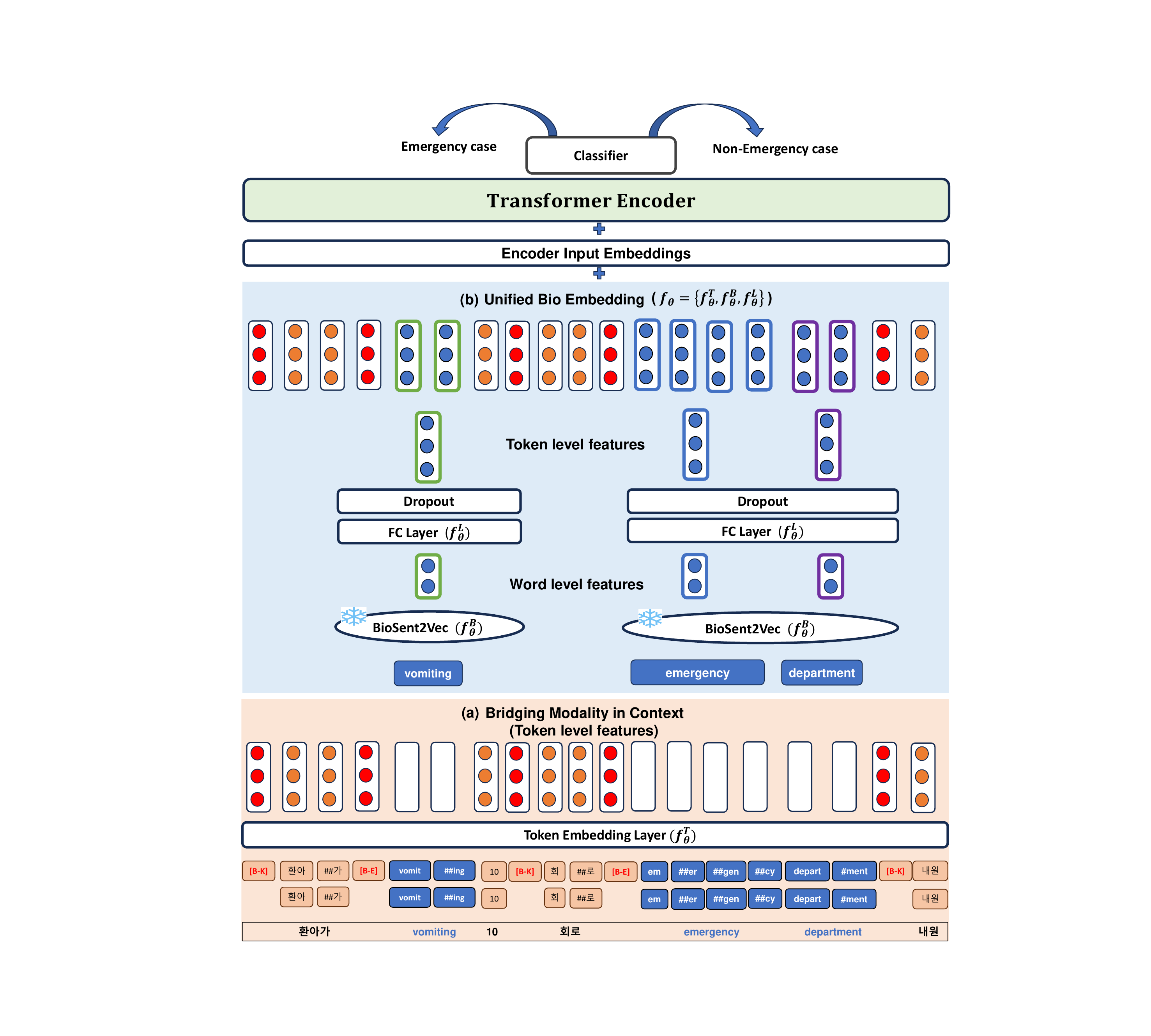}
\vspace{-1.5cm}
\caption{
Overview of BioBridge. 
(a) In the bridging modality in context module, tokenized text input $x^{tok}_{i}$ is processed to reconstruct ${x}_{i}^{bri} =\{\text{[CLS]},\text{[B-K]},\text{[tokens]}^{kor}, \text{[B-E]}, \text{[tokens]}^{eng}, \text{[SEP]}\}$, where modalities for each language are separated using segment tokens. Where ``[B-K]'' and ``[B-E]'' are segment tokens that distinctly identify Korean and English.
In the unified bio-embedding module, English tokens within ${x}_{i}^{bri}$ are restructured at the word level to $\{x^{Eng}_{i}\}^{b}_{i=0}$ which then serve as input for $f^{B}_{\theta}(\{x^{Eng}_{i}\}^{b}_{i=0}) \in \mathbb{R}^{m \times h_{B}}$ to extract medical features. 
These features are subsequently input into ${f}^{L}_{\theta}$, mapping them to the dimension space $h_\mathcal{M}$ of the pre-trained encoder $\mathcal{M}$.
}
\label{figure.3}
\end{center}
\end{figure*}

\subsection{Problem formulation}
Let $X = \{x_{i}\}^{N}_{i=0}$ denote PI, the text input contained in the EMR dataset, and let $X^{Eng} = \{x^{Eng}_{i}\}^{N}_{i=0}$ denote the English text input, $X^{Eng} \subset X$. 
The label $Y = \{y_{i}\}^{N}_{i=0}, y_{i} \in \{0,1\}$ denote Non-emergency cases or Emergency cases.

Tokenized text input for $X$ is defined as $X^{tok} = \{x^{tok}_{i}\}^{N}_{i=0}$.
For instance, if a tokenized input $x^{tok}_{i}$ consists of Korean followed by English, in that order, it is denoted as:
\begin{equation}
    x^{tok}_{i} = \{\text{[CLS]}, \text{[tokens]}^{kor},\text{[tokens]}^{eng},\text{[SEP]}\}
\end{equation}
where, ``[CLS]'' signifies the first token of $x^{tok}_{i}$ and ``[SEP]'' denotes the last token. 
$\text{[tokens]}^{kor}$ denotes the tokens for the tokenized Korean sentence within $x^{tok}_{i}$, and $\text{[tokens]}^{eng}$ denotes the tokens for the tokenized English sentence within $x^{tok}_{i}$.

We denote the batch size as $b$, the total number of tokens within a batch as $n$, and the number of English words as $m$. 
Additionally, we detail three components input to the pre-trained encoder $\mathcal{M}$.

Firstly, the token embedding layer of $\mathcal{M}$, denoted as ${f}^{T}_{\theta}$, is defined as follows:
\begin{equation}
    {f}^{T}_{\theta}(\{x^{tok}_{i}\}_{i=0}^{b})\in \mathbb{R}^{n \times h_\mathcal{M}}
\end{equation}
Where, $h_\mathcal{M}$ is the hidden dimension of $\mathcal{M}$.

Secondly, the medical feature extractor, BioSent2Vec $f^{B}_{\theta}$\cite{chen2019biosentvec}, is defined as:
\begin{equation}
    \label{eq. feature extractor}
    f^{B}_{\theta}(\{x^{Eng}_{i}\}^{b}_{i=0}) \in \mathbb{R}^{m \times h_{B}}
\end{equation}
Where, $h_{B}$ is the hidden dimension of $f^{B}_{\theta}$, and $\theta$ is a fixed parameter.

Thirdly, the Fully Connected (FC) Layer $f^{L}_{\theta}$, designed to map dimensions, is defined as:
\begin{equation}
    \label{eq. Linear mapper}
    \begin{aligned}
        {f}^{L}_{\theta} : \mathbb{R}^{m \times h_{B}}
                        \mathcal{\rightarrow}
                        \mathbb{R}^{m \times h_\mathcal{M}} 
    \end{aligned}
\end{equation}

\textbf{Challenges.} \; 
In practice, applying natural language processing (NLP) to EMRs constructed in non-English-speaking regions presents two primary challenges: 
(1) Code-switched texts between the native language and English are complex and difficult to interpret. Specifically, it is particularly challenging to apply NLP to texts that are code-switched between English and resource-scarce languages, such as Korean.
(2) Due to CS terms predominantly consisting of medical terminology, directly applying models pre-trained in general domains to EMR data proves difficult.

\subsection{BioBridge Framework}
To address the challenges described above, we introduce a two-stage training approach for the BioBridge:
\begin{itemize}
    \item \textbf{Bridging modality in context module}: We add a segment token before each modality within the text to enable the pre-trained encoder $\mathcal{M}$\cite{KoBERT,lee2020kr,devlin2018bert,conneau2019cross,conneau-etal-2020-unsupervised} to effectively distinguish between the modalities (Korean, English) in the given text input $X$.
    \item \textbf{Unified bio-embedding module}: We extract features for English medical words using $f^{B}_{\theta}$\cite{chen2019biosentvec} and unify these features into pre-trained encoder $\mathcal{M}$, enhancing the encoder's applicability to the medical domain.
\end{itemize}
We implement this methodology by training the pre-trained encoder $\mathcal{M}$ using the two training modules. 
Fig.\ref{figure.3} shows an overview of the proposed BioBridge framework.

\subsubsection{Bridging modality in context}
The pre-trained encoder $\mathcal{M}$ must learn to distinguish between Korean and English modalities in the text input $X$.
To facilitate this, we prepend a segment token to each modality in the tokenized input $X^{tok}$.
For instance, by adding segment tokens to $x^{tok}_{i}$, can be denoted as:
\begin{equation}
    \label{eq. CONNECTED MODALITY IN CONTEXT}
    \begin{aligned}
        {x}_{i}^{bri} =  
        \{\text{[CLS]},\text{[B-K]},\text{[tokens]}^{kor}, \text{[B-E]}, \text{[tokens]}^{eng}, \text{[SEP]}\} 
    \end{aligned}    
\end{equation}
Where ``[B-K]'' and ``[B-E]'' are segment tokens that differentiate between Korean and English, respectively.
Additionally, with the inclusion of segment tokens, the dimension of token embedding layer $f^{T}_{\theta}$ is redefined as follows:
\begin{equation}
    \begin{aligned}
        f^{T}_{\theta}(\{{x}_{i}^{bri}\}^{b}_{i=0}) \in \mathbb{R}^{(n+s) \times h_\mathcal{M}}
    \end{aligned}    
\end{equation}
where $s$ denotes the number of added segment tokens.
The related framework is shown in Fig.\ref{figure.3} (a). 

\subsubsection{Unified bio-embedding}
\label{Unified Bio Embedding}
In the unified bio-embedding module, we utilize $f^{B}_{\theta}$ to extract English medical features and integrate them into the pre-trained encoder $\mathcal{M}$. 
The tokenized inputs $\{{x}_{i}^{bri}\}^{b}_{i=0}$ constructed during the bridging modality in context module are segmented at the token level, while $f^{B}_{\theta}$ is optimized for word-level learning.
To address this, we reassemble tokenized English tokens at the token level back into word-level constructs to form $\{{x^{Eng}_{i}\}^{b}_{i=0}}$. 
For example, tokens separated by the tokenizer, such as ``vomit'' and ``\#\#ing,'' are reconstructed into ``vomiting'' as illustrated in Fig.\ref{figure.3} (b).

We then use the word-level reconstructed $\{{x^{Eng}_{i}\}^{b}_{i=0}}$ as input to $f^{B}_{\theta}$, as specified in equation \eqref{eq. feature extractor}, to extract medical features. 
To integrate these extracted medical features into $\mathcal{M}$, the output dimension of $f^{B}_{\theta}$ must be mapped $\mathbb{R}^{h_{B}} \rightarrow \mathbb{R}^{h_\mathcal{W}}$
This mapping is facilitated using $f^{L}_{\theta}$, as described in equation \eqref{eq. Linear mapper}. 

The output from $f^{L}_{\theta}$ produces token-level medical features, which are then integrated into the unified bio-embedding $f_{\theta}$, which can be denoted as:
\begin{equation}
    \begin{aligned}
        f_{\theta}(f^{T}_{\theta},f^{B}_{\theta},f^{L}_{\theta}) \in \mathbb{R}^{m \times h_\mathcal{M}}
    \end{aligned}
\end{equation}
Where, $h_\mathcal{M}$ is the hidden dimension of $\mathcal{M}$.
Ultimately, we integrate unified bio-embedding into the pre-trained encoder $\mathcal{M}$.

\section{Experiments}
In our experiments, we evaluate BioBridge's performance using the proposed EMR dataset. Additionally, all our experiments are designed for a classification task in which the predictive model, such as machine learning-based\cite{chen2016xgboost,cox1958regression,friedman2001greedy,chen2016xgboost,breiman2001random} and pre-trained encoder-based models\cite{KoBERT,lee2020kr,devlin2018bert,conneau2019cross,conneau-etal-2020-unsupervised}, distinguishes between emergency and non-emergency cases\cite{fuchs2016definitions}.

\subsection{Dataset}
\label{sec.3 dataset overview}
The benchmark dataset used in this paper comprises electronic medical records (EMRs) from the Department of Pediatrics at the University Hospital of South Korea.
The study population includes patients under 18 years of age who visited the Department of Pediatrics between January 1, 2012, and December 31, 2021. The EMR dataset includes Present Illness (PI) notes\cite{adler1997history}, urine and blood test results, and records of emergency and non-emergency cases\cite{fuchs2016definitions}. 
Present Illness (PI) is an unstructured free-text clinical note written by a nurse or doctor about the main symptoms observed during the patient's visit to the PED. Additionally, Emergency and Non-emergency cases are data annotated according to whether the patient's visit to the PED was an emergency requiring treatment. Detailed dataset statistics are recorded in Table \ref{Table: Dataset Overview}, and Section IV-B describes the preprocessing of the EMR data. Section\ref{sec : Preprocessing} provides a detailed description of the emergency and non-emergency cases.

\begin{table*}[!ht]
\centering
\setlength{\tabcolsep}{13pt}
\renewcommand{\arraystretch}{1.4}
\begin{tabular}{@{}cc|ccccc@{}}
\toprule
\multicolumn{1}{l}{\textbf{-}} &
  \begin{tabular}[c]{@{}c@{}}Total number \\ of data  \end{tabular} & 
  \begin{tabular}[c]{@{}c@{}}Total number \\ of characters \end{tabular} &
  \begin{tabular}[c]{@{}c@{}}Number of \\ Korean characters \end{tabular} &
  \begin{tabular}[c]{@{}c@{}}Number of \\ English characters \end{tabular} &
  \begin{tabular}[c]{@{}c@{}}Number of \\ Numerical characters \end{tabular} &
  \begin{tabular}[c]{@{}c@{}}Number of \\ special characters \end{tabular}  \\ \midrule
Train    & 56,165     & 16,957,725          & 5,126,952          & 9,396,609           & 671,550          & 1,354,450          \\
Dev      & 14,042     & 4,260,799          & 1,286,444            & 2,362,094           & 166,270            & 339,102         \\
Test     & 17,552   & 5,333,652          & 1,609,512            & 2,959,020           & 211,526            & 426,016      \\
\textbf{total} & \textbf{87,759} & \textbf{26,552,176} & \textbf{8,022,908} & \textbf{14,717,723} & \textbf{1,049,346} & \textbf{2,119,568} \\  \bottomrule
\end{tabular}
\caption{
Statistical Summary of Character Count in Present Illness (PI) texts of the Electronic Medical Record (EMR). 
This table displays the character counts in the PI texts of EMR data, with 8,022,908 Korean and 14,717,723 English characters, indicating that most PI texts are code-switched between Korean and English.
}
\label{Table: Dataset Overview}
\end{table*}

\subsection{Preprocessing of Electronic Medical Record}
\label{sec : Preprocessing}

\begin{figure}[t]
\begin{center}
\includegraphics[scale=0.5]{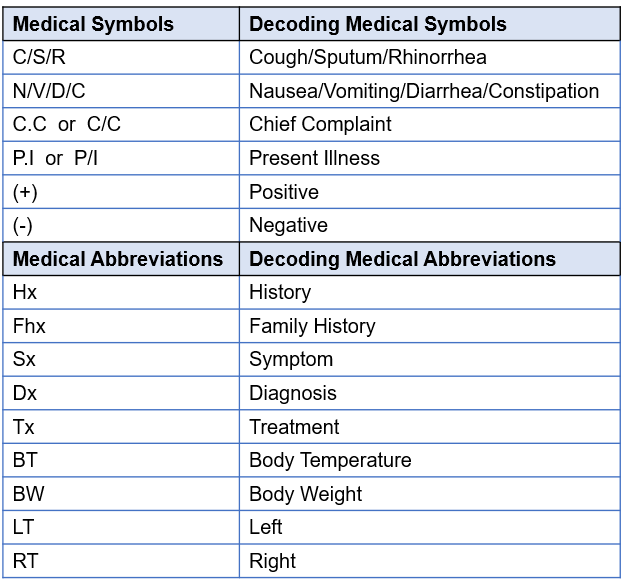} 
\caption{Preprocessing Example of Present Illness (PI) Texts in an Electronic Medical Record (EMR).}
\label{figure.1}
\end{center}
\end{figure}

We constructed an electronic medical record (EMR) dataset of patients who visited Pediatric Emergency Departments (PEDs) by excluding patients with no disposition record and those with missing information. Present Illness (PI) notes contain various medical symbols and abbreviations. For example, in Fig.\ref{figure.1}, the medical symbol ``C/S/R'' stands for ``Cough/Sputum/Rhinorrhea,'' and the abbreviation ``BT'' stands for ``Body Temperature''. We decoded these medical symbols and abbreviations, as shown in Fig.\ref{figure.1}. Additionally, we preprocessed the PI notes by systematically separating Korean, English, numeric values, and special characters when they are followed by a space and introducing a space. We split the preprocessed dataset into training, development, and test sets with ratios of 0.64, 0.16, and 0.20, respectively. In this paper, we integrated PI texts with simple preprocessing into a predictive model.

\subsection{Emergency and Non-emergency cases}
\label{sec: Emergency and Non-emergency cases}
The term “Emergency” refers to an abrupt and often unforeseen occurrence necessitating prompt intervention to reduce potential negative outcomes\cite{fuchs2016definitions}. Cases classified as emergencies encompass patients who received interventions such as blood tests, urinalysis, intravenous hydration, nebulizer treatments, or immediate drug administration in the Pediatric Emergency Department (PED), as well as those who required hospital admission. Conversely, ``Non-emergency'' describes conditions that do not demand urgent medical care, including immediate evaluation, diagnosis, or treatment. Non-emergency cases typically involve patients who were released from the PED without undergoing urgent tests or receiving immediate medication or those who departed with only prescription medication at discharge.

\subsection{Baselines}
\subsubsection{Machine learning based models}
For the Machine Learning (ML) Based Model, we constructed baseline models using one statistical method (Logistic Regression\cite{cox1958regression}) and three tree-based ensemble methods (Gradient Boosting\cite{friedman2001greedy}, XGBoost\cite{chen2016xgboost}, and Random Forest\cite{breiman2001random}).

\textbf{Logistic regression.} \, Logistic regression\cite{cox1958regression} is a statistical model that estimates probabilities using a logistic function widely used for binary classification tasks. It models the probability of a default class (such as an emergency case) based on one or more predictor variables (features).

\textbf{Gradient Boosting.} \, Gradient Boosting\cite{friedman2001greedy} is an ensemble technique that builds models sequentially by correcting the previous models' errors. It uses decision trees as weak learners and focuses on minimizing a loss function, such as mean squared error for regression or log loss for classification, by optimizing the errors of successive models.

\textbf{XGBoost.} \, XGBoost (Advanced Gradient Boosting)\cite{chen2016xgboost} enhances traditional gradient boosting by incorporating advanced features such as regularization to prevent overfitting and a more sophisticated tree pruning and splitting approach.

\textbf{Random Forest.} \, Random Forest\cite{breiman2001random} is an ensemble model consisting of multiple regression trees. It combines several classification trees and trains each on a slightly different set of dataset instances, splitting nodes in each tree considering a limited number of variables. The final predictions of the random forest are made by averaging the predictions of each tree. 

\subsubsection{Encoder based models}
For the encoder base model, we established baselines using both the Korean-specific and the Multilingual-specific encoder-based models.

\textbf{KoBERT.} \, KoBERT\cite{KoBERT} uses the SentencePiece tokenizer and character-level tokenization to reflect and handle the complexity of the Korean language. 
It is trained on Korean Wikidata, which comprises 5 million sentences and 54 million words, with a vocabulary of 8,002 tokens.

\textbf{KR-BERT.} \, KR-BERT\cite{lee2020kr} utilizes the WordPiece tokenizer to capture Korean morphological complexities.
Unlike KoBERT, KR-BERT includes both sub-character and character-level tokenization. For our experiments, we used the KR-BERT-Medium variant. KR-BERT-Medium is an extension of the training data used in the original KR-BERT by adding Korean Wikipedia text, news articles, and Korean online comment datasets. The dataset used for training is 12.37 GB, and the vocabulary size is 20,000.

\textbf{mBERT.} \, Multilingual BERT\cite{devlin2018bert} utilizes Masked Language Modeling (MLM), BERT’s pre-training objective, to train on Wikipedia articles across 104 languages. It uses a shared WordPiece vocabulary comprising 110,000 words to facilitate multilingual processing. This approach enables mBERT to process text in multilingual languages effectively. 

\textbf{XLM.} \, XLM\cite{conneau2019cross} utilizes a cross-lingual pre-training method on a translated parallel corpus and utilizes Byte-Pair Encoding (BPE) for multilingual languages to facilitate cross-lingual transfer learning. This approach has enhanced the performance of mBERT. 

\textbf{XLM-R.} XLM-R\cite{conneau-etal-2020-unsupervised} enhanced the XLM framework by training on 2.5 TB of CommonCrawl data, a significant increase over the dataset used for XLM. This model leverages the robust training approach of RoBERTa\cite{liu2019roberta}, an optimized version of BERT, which includes dynamic masking and removes the next sentence prediction objective. It also supports 100 languages.

\subsection{Performance Metrics}
Our experiments evaluate model performance using several key metrics commonly used in classification tasks: F1 score, Area Under the Receiver Operating Characteristic Curve (AUROC), Area Under the Precision-Recall Curve (AUPRC), and Brier score. To ensure a balanced evaluation, we set the F1 score threshold at 0.595 to align with the label frequency ratio of the label 1 ratio. The F1 score, the harmonic mean of precision and recall, is particularly useful in scenarios with unbalanced datasets. The AUROC measures the model's ability to distinguish between the classes across all thresholds utilizing the False Positive Rate (FPR) and True Positive Rate (TPR). A high AUROC score indicates a strong capacity of the model to differentiate between positive and negative classes, meaning effective model performance.
AUPRC is particularly useful in cases of significant class imbalance, which is the typical dataset. Unlike AUROC, the Area under the precision-recall curve focuses on the performance of the positive class only, making it more sensitive to detecting positive instances. It plots precision and recall for different threshold values, and a higher AUPRC value indicates both high recall and high precision, representing good retrieval of positive cases without many false positives.
The Brier Score is used to measure the accuracy of probabilistic predictions. It is a mean squared error score that quantifies the difference between predicted probabilities and the actual outcomes. A lower Brier score indicates better accuracy, with 0 representing a perfect model and 1 indicating the worst model performance. This metric is particularly useful for evaluating the reliability of a model's probability estimates in our classification task, where predicting the correct probabilities for each class can be more informative than simply predicting the class.

\subsection{Training details}
In our experiments, we developed a model using the PyTorch\cite{paszke2019pytorch} and Transformer\cite{wolf2019huggingface} libraries and utilized a pair of NVIDIA RTX A6000 GPUs for training.

\subsubsection{Training details on machine learning based model}
We used the Term Frequency-Inverse Document Frequency (TF-IDF) to train a Machine Learning (ML) based model. 
TF-IDF is determined by multiplying two metrics: the Term Frequency (TF) score, which reflects the frequency of a word within a document, and the Inverse Document Frequency (IDF) score, which gauges the word's rarity across the entire document corpus. 
A higher TF-IDF score signifies greater importance of the word for prediction. We applied TF-IDF by treating each patient's Present Illnes (PI) as a single document.

\subsubsection{Training details on encoder based model}
\begin{table}[t]
\begin{center}
\setlength{\tabcolsep}{10pt}
\renewcommand{\arraystretch}{1.3}
    \begin{tabular}{lc c}
      \toprule
      \textbf{Model. Param} & Batch Size & Learning Rate \\ 
      \midrule
      \hline
     \multicolumn{3}{c}{\textit{Korean specific encoder based model}} \\
     \hline
      KR-BERT & 64 & 2e-5 \\
      *BioBridge-KR-BERT & 64 & 5e-5 \\
      KoBERT & 64 & 6e-5 \\
      *BioBridge-KoBERT & 64 & 3e-5 \\
      \hline\hline
     \multicolumn{3}{c}{\textit{Multilingual specific encoder based model}} \\
     \hline
      XLM & 22 & 1e-5 \\
      *BioBridge-XLM & 22 & 5e-6 \\
      mBERT\textsubscript{uncased} & 64 & 4e-5 \\
      *BioBridge-mBERT\textsubscript{uncased} & 64 & 1e-5 \\
      mBERT\textsubscript{cased} & 64 & 5e-5 \\
      *BioBridge-mBERT\textsubscript{cased} & 64 & 6e-5\\
      XLM-R\textsubscript{base} & 64 & 5e-5 \\
      *BioBridge-XLM-R\textsubscript{base} & 64 & 3e-5 \\
      \bottomrule
    \end{tabular}
\caption{Hyperparameter used for training Encoder-based model}
\label{table:encoder training details}
\end{center}
\end{table}
In our experiments, we initialized the pre-trained encoder using the checkpoints of KR-BERT, KoBERT, mBERT, XLM, and XLM-R. We also used the ``[CLS]'' token as the final sentence embedding. During training, we saved the checkpoint with the highest f1 score on the development set and evaluated its performance on the test set. Considering the extensive length of most Present Illness (PI) descriptions, we used the maximum sequence length of 512 across all experiments. Also, we used 5 epochs and gridsearch of learning rate $\in \{2e-6,3e-6,5e-6,1e-5,2e-5,3e-5,4e-5,5e-5,6e-5\}$ for all experiments. 
The optimal hyperparameters for the test set are shown in Table \ref{table:encoder training details}. 
Furthermore, all parameters of the medical feature extractor $f^{B}_{\theta}$ in BioBridge are fixed during training.

\subsection{Results}
\begin{table*}[!t]
\setlength{\tabcolsep}{20pt}
\renewcommand{\arraystretch}{1.3}
\begin{center}
\begin{tabular}{lc c c c c}
    \toprule
    \textbf{Model} & \textbf{F1 Score} $\uparrow$ & \textbf{AUROC} $\uparrow$ &\textbf{AUPRC} $\uparrow$ & \textbf{Brier Score} $\downarrow$ \\ 
    \hline \hline
    \multicolumn{5}{c}{\textit{Machine learning based model}} \\
     \hline
    Logistic Regression & 66.60 & 69.58 & 76.24 & 21.43 \\ 
    XGBoost & 66.46 & 69.77 & 76.43 & 21.38 \\
    Gradient Boosting & \underline{66.90} & \underline{70.15} & \underline{76.87}  & \underline{21.25} \\ 
    Random Forest & 66.13 & 67.80 & 74.01 & 22.39 \\
    \hline \hline
     \multicolumn{5}{c}{\textit{Korean specific encoder based model}} \\
     \hline
    KR-BERT & 78.30 & 78.45 & 83.20 & 19.87 \\ 
    *BioBridge-KR-BERT & \textbf{79.40} & \textbf{80.97} & \textbf{85.36} & \textbf{18.92}                      \\
    KoBERT & 78.28 & 80.61 & 85.20 & 18.58 \\
    *BioBridge-KoBERT & \textbf{78.95} & \textbf{80.92} & \textbf{85.27} & \textbf{18.12} \\
    \hline\hline
     \multicolumn{5}{c}{\textit{Multilingual specific encoder based model}} \\
     \hline
    XLM & 78.84 & 79.89 & 84.19 & 21.93 \\
    *BioBridge-XLM & \textbf{79.69} & \textbf{80.64} & \textbf{84.95} & \textbf{18.80} \\
    mBERT\textsubscript{uncased} & 79.48 & 81.47 & 85.82 & 18.36 \\
    *BioBridge-mBERT\textsubscript{uncased} & \textbf{79.50} & \textbf{81.98} & \textbf{86.22} & \textbf{17.43} \\
    mBERT\textsubscript{cased} & 79.89 & 81.18 & 85.40  & 19.17 \\
    *BioBridge-mBERT\textsubscript{cased} & \textbf{79.98} & \textbf{81.76} & \textbf{85.97} & \textbf{18.42} \\
    XLM-R\textsubscript{base} & 79.55 & 81.05 & 84.07 & \textbf{18.78} \\
    *BioBridge-XLM-R\textsubscript{base} & \textbf{79.99} & \textbf{81.32} & \textbf{85.57} & 19.59   \\
    \bottomrule
\end{tabular}
\caption{
Performance Comparison of Machine learning-Based and encoder-based models in classifying emergency cases.
*: results from the proposed BioBridge experiments. 
The best result within each row block is shown in bold, and The best results from the Machine learning-based model are underlined.
}
\label{main resutls}
\end{center}
\end{table*}

Table \ref{main resutls} shows the evaluation results on the test dataset. The comparison between machine learning-based models and encoder-based models (Korean and multilingual encoder-based models) clearly shows the superiority of encoder-based models. For the Machine Learning models, while Gradient Boosting achieved the highest F1 score of 66.90, the KR-BERT significantly improved, reaching an F1 score of 78.30. The improvement extends to AUROC, AUPRC, and Brier score metrics, revealing that Korean encoder-based models not only boost prediction accuracy but also refine the calibration of those predictions. The ML-based model was trained based on TF-IDF, so it would have performed worse than the encoder-based model that considered the context.

In the category of Korean-specific encoder-based models, *BioBridge-KR-BERT significantly outperforms the baseline KR-BERT across all evaluated metrics. The F1 score improves by 1.1\%, AUROC by 2.52\%, AUPRC by 2.16\%, and the Brier score decreases by 0.95\%. These results indicate that *BioBridge-KR-BERT enhances classification performance across various thresholds. Moreover, BioBridge demonstrates superiority in the accuracy of probabilistic predictions. This trend aligns with the performance gains observed between BioBridge-KoBERT and the base KoBERT model. Demonstrate the efficacy of BioBridge with Korean-based encoder models.

For the multilingual-specific encoder-based model, *BioBridge-XLM showed an increase of 0.85\% in the F1 score, 0.75\% in AUROC, and 0.76\% in AUPRC compared to the baseline XLM. Notably, the Brier score decreased by 3.04\%, signifying that BioBridge enhanced the prediction accuracy and the reliability and calibration of the model’s outputs. Moreover, *BioBridge-mBERT\textsubscript{uncased} achieved the lowest Brier score.

BioBridge's consistent enhancement in performance across various encoder-based models attests to its robustness and adaptability to different encoder architectures and languages. These outcomes endorse the BioBridge method as a powerful strategy for classification tasks in the medical domain, notably in settings dealing with bilingual EMRs.

\section{Ablation study}
In this section, we perform extensive ablation studies to support our model design. For all experiments, we establish a multilingual-based model as our baseline. The BioBridge framework was designed in two modules. First the ``bridging modality in context'' module, which bridges modalities (Korean, English). This module is essential due to it connects each modality (Korean, English) in the CS context with a token. Excluding the token that bridges each modality in the CS sentence becomes a vanilla encoder-based model. The second module, ``unified bio-embedding,'' aims to infuse the encoder-based model, which lacks medical domain knowledge, with expertise in English medical terminology. The ``unified bio-embedding'' module extracts medical features from the medical feature extractor $f^{B}_{\theta}$ and incorporates them into the encoder base model. The vanilla encoder-based model is obtained by excluding the ``unified bio-embedding'' module. The optimal hyperparameters for the test set are shown in Table \ref{hyperparam_ablation}.

\begin{table}[t]
\setlength{\tabcolsep}{10pt}
\renewcommand{\arraystretch}{1.3}
\begin{center}
    \begin{tabular}{lc c}
      \toprule
      \textbf{Model. Param} & Batch Size & Learning Rate \\ 
      \midrule
      \hline\hline
     \multicolumn{3}{c}{\textit{Multilingual specific encoder based model}} \\
     \hline
      XLM &  &  \\
      \; w/ Bridging modality & 22 & 1e-5 \\
      \; w/ Bio-embedding & 22 & 5e-6 \\
      mBERT\textsubscript{uncased} & & \\
      \; w/ Bridging modality & 64 & 4e-5 \\
      \; w/ Bio-embedding & 64 & 4e-5 \\
      mBERT\textsubscript{cased} & & \\
      \; w/ Bridging modality & 64 & 5e-5 \\
      \; w/ Bio-embedding & 64 & 4e-5 \\
      XLM-R\textsubscript{base} & & \\
      \; w/ Bridging modality & 64 & 3e-5 \\
      \; w/ Bio-embedding & 64 & 3e-5 \\
      \bottomrule
    \end{tabular}
\caption{Hyperparameter used for ablation studies on Multilingual Specific Encoder based model}
\label{hyperparam_ablation}
\end{center}
\end{table}

\subsection{Results}
We show insightful results for the BioBridge framework through extensive ablation studies. The results of the ablation study are shown in Table \ref{ablation_table}. Additionally, The optimal hyperparameters for the test set are shown in Table \ref{hyperparam_ablation}.

\textbf{Bridging modality in context.} \;
Integrating the ``bridging modality in context'' module into our baseline multilingual encoder-based model led to notable improvements across various performance metrics. In XLM, F1 score increased by 0.26\%, AUROC increased by 0.3\%, AUPRC increased by 0.33\%, and Brier score decreased by 0.17\%. These results highlight the importance of this module in enhancing comprehension of the context within code-switched scenarios.

\textbf{Uninifed bio-embedding.} \;
Adding the ``unified bio-embedding'' module to our baseline multilingual encoder-based model also consistently enhanced performance on almost all metrics. In mBERT\textsubscript{cased}, we observed a 0.62\% decrease in the F1 score but a 0.25\% increase in AUROC and a 0.57\% increase in AUPRC, demonstrating robust performance at all thresholds. Notably, the Brier score decreased by 1.34\%, affirming enhanced reliability and accuracy in probabilistic predictions. This improvement demonstrates that the ``unified bio-embedding'' module can effectively inject medical knowledge into a multilingual-based model trained on a general domain.

The results of these ablation studies highlight the importance of both modules in improving the encoder base model's ability to understand bilingual EMR accurately.

\begin{table}[t]
\renewcommand{\arraystretch}{1.3}
\begin{center}
\resizebox{\columnwidth}{!}{%
\begin{tabular}{lc c c c c}
    \toprule
    \textbf{Model} & \textbf{F1} $\uparrow$ & \textbf{AUROC}$\uparrow$ &\textbf{AUPRC}$\uparrow$ &\textbf{Brier}$\downarrow$\\
    \hline \hline
    \multicolumn{5}{c}{\textit{Multilingual specific encoder based model}} \\
    \hline
    XLM & 78.84 & 79.89 & 84.19 & 21.93 \\
    \; w/ Bridging modality & 79.10 & 80.19 & 84.52 & 21.76 \\
    \; w/ Bio-embedding & 79.20 & 80.02 & 84.39 & 19.00\\
    *BioBridge-XLM & \textbf{79.69} & \textbf{80.64} & \textbf{84.95} & \textbf{18.80} \\
    \hline
    
    mBERT\textsubscript{cased} & 79.89 & 81.18 & 85.40  & 19.17 \\
    \; w/ Bridging modality & \textbf{80.08} & 81.36 & 85.75  & 19.36 \\
    \; w/ Bio-embedding & 79.27 & 81.43 & 85.97 & \textbf{17.83} \\
    *BioBridge-mBERT\textsubscript{cased} & 79.98 & \textbf{81.76} & \textbf{85.99} & 18.42 \\
    \hline
    
    mBERT\textsubscript{uncased} & 79.48 & 81.47 & 85.82 & 18.36 \\
    \; w/ Bridging modality & 78.80 & 81.57 & 86.00 & 17.96 \\
    \; w/ Bio-embedding & 78.69 & 81.34 & 85.66 & 17.49 \\
    *BioBridge-mBERT\textsubscript{uncased}& \textbf{79.50} & \textbf{81.98} & \textbf{86.22} & \textbf{17.43}  \\
    \hline

    XLM-R\textsubscript{base} & 79.55 & 81.05 & 84.07 & 18.78 \\
    \; w/ Bridging modality & 79.27 & 81.32 & 85.48 & \textbf{18.75} \\
    \; w/ Bio-embedding & 79.60 & \textbf{81.67} & \textbf{85.99} & 23.33 \\
    *BioBridge-XLM-R\textsubscript{base} & \textbf{79.99} & 81.32 & 85.57 & 19.59 \\
    
    \bottomrule
\end{tabular}
}
\end{center}
\caption{
Ablation Study of a multilingual encoder-Based Model for Emergency Case Classification.
*: results from the proposed BioBridge experiments. 
The best result within each row block is shown in bold.
}
\label{ablation_table}
\end{table}

\section{Limitation}
This paper proposes applying Natural Language Processing (NLP) to Electronic Medical Records (EMRs) to improve decision-making processes in pediatric emergency departments (PEDs). Our work has three primary limitations. Firstly, data privacy and security present challenges. EMRs contain sensitive personal information, necessitating strict compliance with privacy laws and regulations, which restricts the free sharing of data and limits its broader application. Secondly, the quality and consistency challenges. EMR data is written in free text, which makes the data inconsistent. Therefore, data input in different formats may hinder the model's ability to learn uniformly, potentially affecting the generalizability of the results. Thirdly, Model Interpretability challenges. The complexity of the Transformer model architecture used in this work reduces its interpretability. In a healthcare setting, where decisions can have significant implications for patient outcomes, the inability to fully explain how model predictions are made can severely hinder trust and adoption. 

In future work, we plan to increase the generalization performance of our model in healthcare settings by obtaining EMR datasets from more hospitals while maintaining compliance.
Furthermore, while this paper focuses on Korean-English code-switched EMRs, we intend to extend our approach to include code-switched EMRs from other languages with English to explore its broader applicability in multilingual healthcare environments.
Additionally, to make the model more transparent and interpretable, we plan to use methods such as SHapley Additive exPlanations (SHAP)\cite{lundberg2017unified} to provide an interpretation of the model.

\section{Conclusion}
In this paper, we proposed BioBridge, a framework that unified bio-embedding with bridging modality in code-switched EMR. This approach achieved state-of-the-art performance in classifying emergency cases in Korean-English code-switched EMRs by fine-tuning a transformer encoder-based model. This approach addresses the critical challenges of overcrowding in PEDs and pioneers using code-switched EMRs with multilingual NLP techniques, setting a new standard for future applications.

\bibliographystyle{unsrt}
\bibliography{reference}

\EOD

\end{document}